\documentclass[10pt,twocolumn,letterpaper]{article}

\usepackage{cvpr}
\usepackage{times}
\usepackage{epsfig}
\usepackage{graphicx}
\usepackage{amsmath}
\usepackage{amssymb}

\usepackage{breqn}
\usepackage{tabularx}
\usepackage{multirow}
\usepackage{makecell}
\usepackage{lineno}

\usepackage{epstopdf}
\usepackage{subfigure}
\usepackage{color}

\usepackage{epstopdf}
\usepackage{subfigure}
\usepackage{color}
\usepackage{caption}
\usepackage[american]{babel}
\usepackage{cite}
\usepackage[american]{babel}
\usepackage{bbding}
\usepackage{cuted}
\usepackage[pagebackref=true,breaklinks=true,letterpaper=true,colorlinks,bookmarks=false]{hyperref}

 \cvprfinalcopy % *** Uncomment this line for the final submission

\def\cvprPaperID{***} % *** Enter the CVPR Paper ID here

\ifcvprfinal\fi
\begin{document}

%%%%%%%%% TITLE
\title{Learning Symmetry Consistent Deep CNNs for Face Completion}

\author{Xiaoming Li$^{1}$, Ming Liu$^{1}$, Jieru Zhu$^{1}$, Wangmeng Zuo$^{1(}$\Envelope$^)$, Meng Wang$^{2}$, Guosheng Hu$^{3}$, Lei Zhang$^{4}$\\
\small $^1$Harbin Institute of Technology,
$^2$Hefei University of Technology,
$^3$Anyvision, $^4$The Hong Kong Polytechnic University\\
{\tt\small \{csxmli,wmzuo\}@hit.edu.cn}
}
\twocolumn[{
 \renewcommand\twocolumn[1][]{#1}
 \maketitle
 {
  \centering
  \vspace{-7mm}
  \setlength{\tabcolsep}{0.2mm}
{
		\begin{tabular}{cccccc}
		\includegraphics[width=0.165\linewidth]{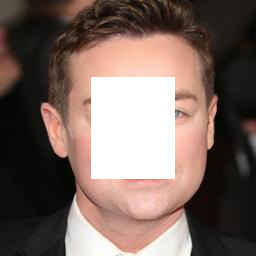}&
		\includegraphics[width=0.165\linewidth]{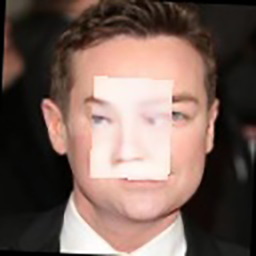} &
		\includegraphics[width=0.165\linewidth]{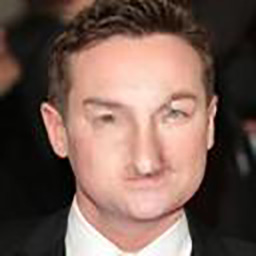} &
		\includegraphics[width=0.165\linewidth]{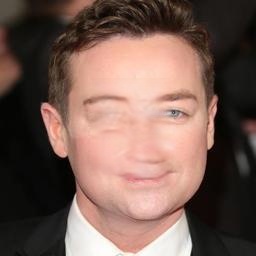} &
		\includegraphics[width=0.165\linewidth]{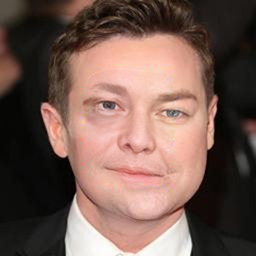} &
		\includegraphics[width=0.165\linewidth]{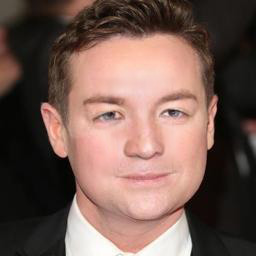} \\
		\includegraphics[width=0.165\linewidth]{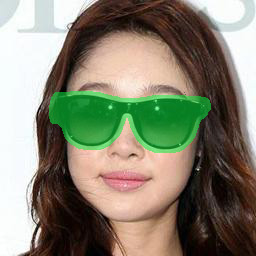} &
		\includegraphics[width=0.165\linewidth]{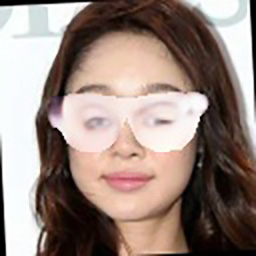} &
		\includegraphics[width=0.165\linewidth]{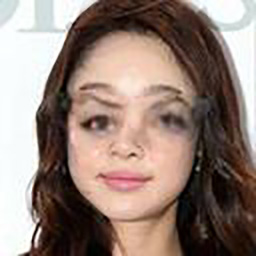} &
		\includegraphics[width=0.165\linewidth]{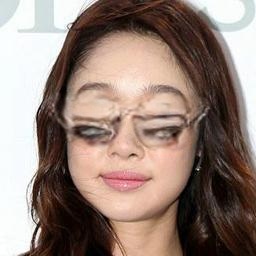} &
		\includegraphics[width=0.165\linewidth]{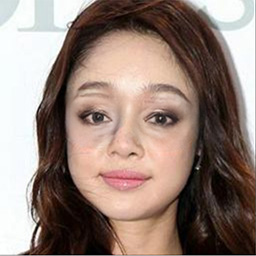} &
		\includegraphics[width=0.165\linewidth]{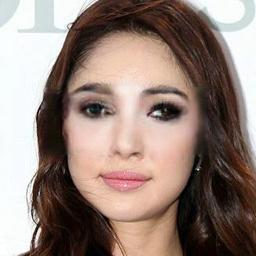} \\
		(a) Input & (b) Iizuka \emph{et al.}~\cite{iizuka2017globally} &  (c) Li \emph{et al.}~\cite{GFC2017CVPR} &
		(d) Yu \emph{et al.}~\cite{Yu2018CVPR} & (e) Liu \emph{et al.}~\cite{liu2018image} &(f) Ours\\
		\end{tabular}
}
\captionof{figure}{
   Completion results of face images with synthetic and real occlusions.
   \label{fig:symmetry}
  }
		%\end{center}
 }
 \vspace{10pt}
}]

%%%%%%%%% ABSTRACT

\begin{abstract}
\vspace{-9pt}
Deep convolutional networks (CNNs) have achieved great success in face completion to generate plausible facial structures.
These methods, however, are limited in maintaining global consistency among face components and recovering fine facial details.
On the other hand, reflectional symmetry is a prominent property of face image and benefits face recognition and consistency modeling, yet remaining uninvestigated in deep face completion.
In this work, we leverage two kinds of symmetry-enforcing subnets to form a symmetry-consistent CNN model (i.e., SymmFCNet) for effective face completion.
For missing pixels on only one of the half-faces, an illumination-reweighted warping subnet is developed to guide the warping and illumination reweighting of the other half-face.
As for missing pixels on both of half-faces, we present a generative reconstruction subnet together with a perceptual symmetry loss to enforce symmetry consistency of recovered structures.
The SymmFCNet is constructed by stacking generative reconstruction subnet upon illumination-reweighted warping subnet, and can be end-to-end learned from training set of unaligned face images.
Experiments show that SymmFCNet can generate high quality results on images with synthetic and real occlusion, and performs favorably against state-of-the-arts.
\end{abstract}

%%%%%%%%% BODY TEXT
\vspace{-23pt}
\section{Introduction}
%
%%%%%%%%%%%%%%%%%%%%%%%%%%%%%%%
\begin{figure*}
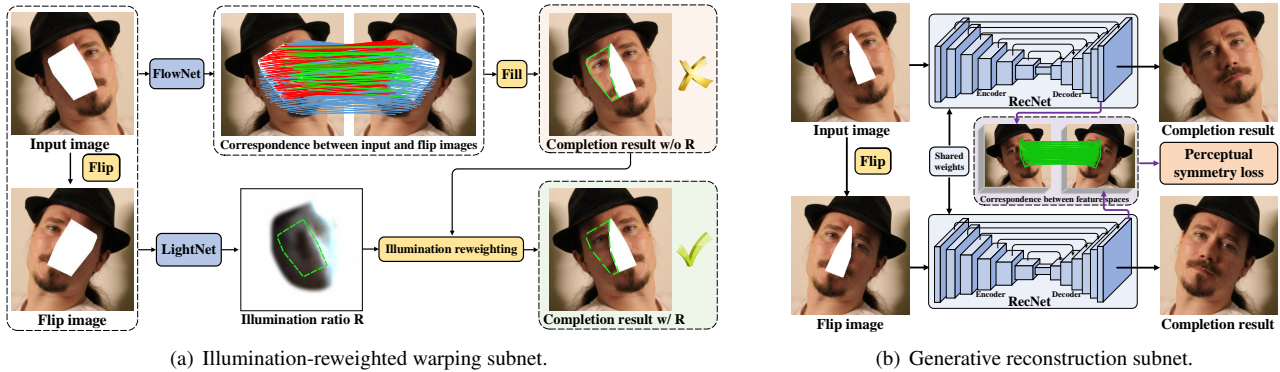
 \centering
\subfigure[Illumination-reweighted warping subnet.] { \label{fig:fig2a}
\includegraphics[width=1.15\columnwidth,height = 0.531\columnwidth]{./eps/Architecture/Fig2a.pdf}
}
\hspace{1.5em}
\subfigure[Generative reconstruction subnet.] { \label{fig:fig2b}
\includegraphics[width=0.784\columnwidth,height = 0.531\columnwidth]{./eps/Architecture/Fig2b.pdf}
}
\caption{\small Overview of our SymmFCNet. {\color{red}Red}, {\color{green}green} and {\color{blue}blue} lines represent the pixel-wise correspondence between the input and the flip image. {\color{red}Red:} missing pixels (input) to non-occluded pixels (flip); {\color{green}Green:} missing pixels (input) to  missing pixels (flip); {\color{blue}Blue:} remaining pixels (input) to remaining pixels (flip).
}
\label{fig}
\end{figure*}
The task of face completion is to fill in missing facial pixels with visually plausible hypothesis~\cite{yeh2017semantic,GFC2017CVPR}.
The generated solutions for missing parts aim to restore semantic facial structures and realistic fine details, but are not required to exactly approximate the unique ground-truth.
Unlike images of natural scene, face images usually contain little repetitive structures~\cite{Yu2018CVPR}, which further increases the difficulties of face completion.
Moreover, face completion can also be used in many real world face-related applications such as unwanted content removal (\emph{e.g.}, glasses, scarf, and HMD in interactive AR/VR), interactive face editing, and occluded face recognition.

Recently, along with the development of deep learning, significant progress has been made in image inpainting~\cite{Yu2018CVPR,Song2017Contextual,yan2018shift,pathak2016context,yang2017high} and face completion~\cite{GFC2017CVPR,iizuka2017globally,Yu2018CVPR,liu2018image}.
The existing methods generally adopt the generative adversarial network (GAN)~\cite{goodfellow2014generative} framework which involves a generator and a discriminator.
On one hand, contextual attention~\cite{Yu2018CVPR} and shift-connection~\cite{Song2017Contextual,yan2018shift} have been introduced into the baseline generator (\emph{i.e.}, context-encoder)~\cite{pathak2016context} to exploit surrounding repetitive structures for generating visually plausible content with fine details.
On the other hand, global and local discriminators are incorporated to obtain globally consistent result with locally realistic details~\cite{iizuka2017globally,GFC2017CVPR}, and semantic parsing loss is also adopted to enhance the consistency of face completion result~\cite{GFC2017CVPR}.

However, face completion is not a simple application of image inpainting, and remains not well solved.
Fig.~\ref{fig:symmetry} illustrates the results by state-of-the-art CNN-based methods, including Iizuka \etal~\cite{iizuka2017globally}, Li \etal~\cite{GFC2017CVPR}, Yu \etal~\cite{Yu2018CVPR}, and Liu \etal~\cite{liu2018image}.
Because face images are generally
%Due to that face images generally are
of non-repetitive structures, blurry results remain inevitable in the methods \cite{GFC2017CVPR,iizuka2017globally,Yu2018CVPR} based on auto-encoder and contextual attention.
Furthermore, from the top image in Fig.~\ref{fig:symmetry}(e),
%even
although the generated right eye by~\cite{liu2018image} is locally satisfying, it is globally inconsistent with the left eye.

In this work, we present a deep symmetry-consistent face completion network (SymmFCNet), which leverages face symmetry to improve the global consistency and local fine details of face completion result.
%
%The reflectional symmetry of face images is well known and has been adopted to face recognition and consistency modeling.
%
{The reflectional symmetry of face images, which has been widely adopted to face recognition and consistency modeling~\cite{song2006face,harguess2011cvpr,dovgard2004statistical},} remains a non-trivial issue to exploit the symmetry property in face completion due to the effect of illumination and pose.
Non-symmetric lighting causes the pixel value not equal to the corresponding pixel value in the other half-face.
The deviation from frontal face further breaks the reflectional symmetry and makes the pixel correspondence between two half-faces much more complicated.

As shown in Fig.~\ref{fig}, the correspondence between two half-faces can be divided into three types: %The first one is that
(1) The missing pixels in input image correspond to non-occluded pixels in flip image ({\color{red}red lines} in Fig.~\ref{fig:fig2a}), which indicates that these missing pixels can be filled by their symmetric ones.
%The second one is that
(2) The missing pixels in input image correspond to missing pixels in flip image ({\color{green}green lines} in Fig.~\ref{fig:fig2a}), which indicates that these missing pixels can only be filled by generation.
%The last one is that
(3) The remaining pixels in input image correspond to other pixels in flip image ({\color{blue}blue lines} in Fig.~\ref{fig:fig2a}). %With this in mind,
Based on this, we present two mechanisms to leverage symmetric consistency for filling in two types of missing pixels.

For missing pixels happened on only one of the half-faces (see {Fig.~\ref{fig:fig2a}}), it is natural to fill them by reweighting the illumination of the corresponding pixels in the other half-face {(the red correspondence in Fig.~\ref{fig:fig2a})}.
To cope with pose and illumination variation between half-faces, we suggest an illumination-reweighted warping subnet of two parts: (i) a FlowNet to establish the correspondence map between two half-faces, and (ii) a LightNet to indicate the ratio of illumination between two half-faces.
For missing pixels happened on both of the half-faces (see {Fig.~\ref{fig:fig2b}}), perceptual symmetry loss is incorporated with a generative reconstruction subnet (RecNet) for symmetry-consistent completion.
Based on the correspondence map established by FlowNet, the perceptual symmetry loss is defined on the decoder feature layer to alleviate the effect of illumination inconsistency
%variation
between two half-faces.
To sum up, our full SymmFCNet can be constructed by stacking generative reconstruction subnet upon illumination-reweighted warping subnet.
Perceptual symmetry, reconstruction and adversarial losses are deployed on RecNet to end-to-end train the full SymmFCNet.
While illumination consistency loss, landmark loss and total variation (TV) regularization are employed to illumination-reweighted warping subnet for improving the training stability of FlowNet and LightNet.

%Extensive experiments are conducted to evaluate our SymmFCNet.
%
Experiments show that illumination-reweighted warping is effective in filling in missing pixels happened on only one of the half-faces.
In contrast, the RecNet can not only generate symmetry-consistent result for missing pixels happened on both of the half-faces, but also benefit the refinement of the result by illumination-reweighted warping.
In terms of quantitative metrics and visual quality, our SymmFCNet performs favorably against state-of-the-arts~\cite{GFC2017CVPR, iizuka2017globally,Yu2018CVPR,liu2018image}, and achieves high quality results on face images with real occlusions.
The contributions of this work include:
\begin{itemize}
	\item
	An illumination-reweighted warping network for filling in missing pixels on only one of the half-faces.
    \item A generative reconstruction network equipped with perceptual symmetry loss for the inpainting of missing pixels on both of the half-faces.
	\item Our full SymmFCNet for high quality symmetry-consistent face completion with either rectangular or irregular missing regions.
	\item Favorable performance of SymmFCNet in comparison to state-of-the-arts~\cite{iizuka2017globally,GFC2017CVPR,Yu2018CVPR,liu2018image}.
\end{itemize}
%-------------------------------------------------------------------------
\section{Related Work}\label{relatedwork}
In this section, we briefly review the relevant work of three sub-fields: deep image inpainting, deep face completion and the applications of symmetry in face analysis.

\textbf{Deep Image Inpainting.} Image inpainting aims to fill in missing pixels in a seamless manner~\cite{bertalmio2000image}, which has wide applications, such as restoration of damaged image and unwanted content removal.
Recently, motivated by the unprecedented success of GAN in many vision tasks like style transfer~\cite{li2016precomputed,huang2017arbitrary}, image-to-image translation~\cite{pix2pix2016,CycleGAN2017}, image super-resolution~\cite{ledig2017photo} and face attribute manipulation~\cite{li2016deep}, deep CNNs have also greatly facilitated the development of image inpainting.
Originally, Pathak \emph{et al.}~\cite{pathak2016context} present an encoder-decoder (\emph{i.e}., context encoder) network to learn the image semantic structure for the recovery of the missing pixels, and an adversarial loss is deployed to enhance the visual quality of the inpainting result.
Subsequently, global and local discriminators~\cite{iizuka2017globally} are adopted for better discrimination between real images and inpainting results.
%
%While
In addition, dilated convolution~\cite{iizuka2017globally} and partial convolution~\cite{liu2018image} are introduced to improve the training of generator.
To exploit the repetitive structures in surrounding contexts, multi-scale neural patch synthesis (MNPS)~\cite{yang2017high} is suggested, and contextual attention~\cite{Yu2018CVPR} and shift-connection~\cite{yan2018shift,Song2017Contextual} are further presented to overcome the inefficiency of MNPS.
%
%In comparison,
Unlike natural images, face images generally exhibit non-repetitive structures and are more sensitive to semantic consistency and visual artifacts, making it difficult to directly apply general-purposed inpainting models.
%, making it difficult to address by directly applying general-purposed inpainting models.

%
\textbf{Deep Face Completion.}
{Apart from the aforementioned image inpainting methods,}
%Besides these general image completion methods mentioned above, as for face completion,
Yeh \emph{et al.}~\cite{yeh2017semantic} develop a semantic %image inpainting
{face completion} method, which exploits the trained GAN to find the closest encoding and then fill the missing pixels by considering both context discriminator and corrupted input image.
%
%Similar to~\cite{iizuka2017globally} (\hu{general image inpainting}),
%
Li \emph{et al.}~\cite{GFC2017CVPR} learn a generative model to recover missing pixels by minimizing the combination of reconstruction loss, local and global adversarial losses as well as semantic parsing loss.
For better recovery of facial details, Zhao \emph{et al.}~\cite{zhao2018identity} suggest a guidance image from the extra non-occluded face image to facilitate identity-aware completion.
However, the introduction of guidance image certainly limits its wide applications, and its performance degrades remarkably when the guidance and occluded images are of different poses.
Instead of guidance image, we leverage the symmetry of face images to establish the correspondence between two half-faces, which is then used to guide the generation of high quality completion result.

\textbf{Face Symmetry.}
Symmetry is closely related to the human perception, understanding and discovery of images, and also has received upsurging interests in computer vision~\cite{liu2010computational,funk20172017,dovgard2004statistical,passalis2011using,song2006face,harguess2011cvpr}.
In computational symmetry, numerous methods have been proposed to detect reflection, rotation, translation and medial-axis-like symmetries from images~\cite{liu2010computational,funk20172017}.
Reflectional symmetry is also an important characteristic of face images, which has been used to assist 3D face reconstruction~\cite{dovgard2004statistical}, 3D face alignment~\cite{passalis2011using} and face recognition~\cite{song2006face,harguess2011cvpr}.
In addition, Huang \emph{et al.}~\cite{huang2017beyond} adopt symmetry loss on pixel and Laplacian space for identity-preserving face frontalization.
%
%Different from
Unlike~\cite{huang2017beyond}, we present a more general scheme for modeling face symmetry for face completion.

\section{Method}\label{method}

Face completion aims at learning a mapping from occluded face $I^o$ as well as its binary indicator mask $M$ to a desired completion result $\hat{I}$ (\emph{i.e.}, an estimation of the ground-truth ${I}$).
Here, the images $I^o$, $M$, and $\hat{I}$ are of the same size $h \times w$, and $M(i,j) = 0$ indicates the pixel at $(i,j)$ is missing.
To exploit face symmetry, we present our two-stage SymmFCNet to generate symmetry-consistent completion result.
In the first stage, an illumination-reweighted warping subnet is deployed to fill in missing pixels happened on only one of the half-faces (see Fig.~\ref{fig:fig2a}), where a FlowNet is included to establish the correspondence between two half-faces.
%, and illumination consistency and landmark losses are introduced to train the subnet.
%
In the second stage, a generative reconstruction subnet is used to handle missing pixels happened on both of the half-faces and further refine the inpainting result (see Fig.~\ref{fig:fig2b}).
Using the output of FlowNet, we define a perceptual symmetry loss on the decoder feature layer to enforce symmetry consistent completion.
%
%Reconstruction and adversarial losses are also incorporated for the end-to-end training of SymmFCNet.
%
In this section, we first detail the architecture of SymmFCNet and then define the learning objective.

\subsection{Illumination-reweighted warping}\label{sec::wholeNet}

Unlike general-purposed image inpainting, face is a highly structured object with prominent reflectional symmetric characteristic.
Thus, when the missing pixels are within only half of the face, it is reasonable to fill them based on the corresponding pixels in the other half-face.
To this end, we should solve the illumination inconsistence and create correspondence
%address both the correspondence and the illumination variation
between the pixels from two half-faces.
For example, given a missing pixel $(i,j)$, if its corresponding pixel $(i^{\prime},j^{\prime})$ in the other half-face and their illumination ratio $R(i,j) = \frac{I(i,j)}{I(i^{\prime},j^{\prime})}$ is known, the value $\hat{I}(i,j)$ can then be computed by $I^o(i^{\prime},j^{\prime}) R(i,j)$ (Note that $I^o(i^{\prime},j^{\prime}) = I(i^{\prime},j^{\prime}) $).
In the following, we introduce a FlowNet and a LightNet for computing pixel correspondence and illumination ratio, respectively.

\subsubsection{FlowNet}
One may establish the correspondence between the pixels from two half-faces by direct matching.
However such approach is computational costly and the annotation of dense correspondence is also practically infeasible.
Instead, we introduce the flip image $I^{o'}$ ($M^{'}$) of occluded face (mask) $I^o$ ($M^{}$), and adopt a FlowNet which takes both $I^{o'}$ and $I^o$ to predict the flow field $\Phi = (\Phi^x,\Phi^y)$,
\begin{equation}
\label{eq:flow}
\Phi=\mathcal{F}_w(I^o, I^{o'}; \Theta_w),
\end{equation}
where $\Theta_w$ denotes the FlowNet model parameters.
Given a pixel $(i,j)$ in $I^o$, $(\Phi^x_{i,j},\Phi^y_{i,j})$  indicates the position of its corresponding pixel in $I^{o'}$.
Note that $I^{o'}$ is the flip image of $I^o$.
Thus, $I^o(i,j)$ and $I^{o'}(\Phi^x_{i,j},\Phi^y_{i,j})$ are a pair of corresponding pixels from different half-faces, and the correspondence between two half-faces is then constructed.

With $\Phi$, the pixel value at $(i,j)$ of the warped image $I^w$ is defined as the pixel value at $(\Phi^x_{i,j},\Phi^y_{i,j})$ of the flipped image $I^{o'}$. Since $\Phi^x_{i,j}$ and $\Phi^y_{i,j}$ are real numbers, then $I^{o'}_{ (\Phi^x_{i,j},\Phi^y_{i,j})}$ can be bilinear interpolated by its 4 surrounding neighboring pixels. Thus, the warped image $I^w_{i,j}$ can be computed as the interpolation result:
\begin{equation}
\label{eq:phi}
%\scriptsize
\small
I^{w}_{i,j} = \sum_{(h,w) \in \mathcal{N}} I^{o'}_{h,w} \max(0, 1 - |\Phi^{y}_{i,j} - h|) \max(0, 1 - |\Phi^{x}_{i,j} - w|),
\end{equation}
where $\mathcal{N}$ denotes the 4-pixel neighbors of $(\Phi^x_{i,j},\Phi^y_{i,j})$.
Analogously, the warped mask image $M^{w}$ of $M^{'}$ can be given by:
\begin{equation}
\label{eq:phi}
%\scriptsize
\small
M^{w}_{i,j} = \sum_{(h,w) \in \mathcal{N}} M^{'}_{h,w} \max(0, 1 - |\Phi^{y}_{i,j} - h|) \max(0, 1 - |\Phi^{x}_{i,j} - w|).
\end{equation}
By defining $M^{s1} = M^{w} \odot (1-M)$, we can then identify the missing pixels $(i, j)$ within only half of the face as $M^{s1}_{i,j} = 1$. Here, $\odot$ represents the element-wise product operation.

The FlowNet adopts the encoder-decoder architecture which is the same as pix2pix~\cite{pix2pix2016} except that the inputs contain 6 channels rather than 3 ones.
%
%The input encoder consists of 8 convolution layers with kernel size $4 \times$ and stride $2$.
%
%And the decoder takes 8 deconvolution layers to predict the 2 channels dense flow field $\Phi$.
%
%All of the layers except the first layer in encoder and last layer in decoder adopt the Convolution-BatchNorm-ReLU form.
%
As for the last activation function, we employ \emph{tanh} to normalize the two channels coordinates to the range $[- 1,1]$.
Please refer to the appendix for more details of FlowNet.

%Due to
Because it is unpractical to annotate the dense correspondence between left and right half-faces, alternative losses are required to train FlowNet.
In~\cite{ganin2016deepwarp,yeh2016semantic,zhou2016view}, the losses are enforced on the warped images.
%
%But f
For face completion, however, the ground-truth of warped image is unknown, and the two half-faces may be of different illumination, making it unsuitable to use $I^{o}$ as the ground-truth.

Following~\cite{li2018learning}, we train FlowNet in a semi-supervised manner by incorporating landmark loss with a TV regularizer.
Given the ground-truth image $I$, we detect its 68 facial landmarks $\left\{(x^g_i,y^g_i)\left| {_{i = 1}^{68}} \right.\right\}$ through~\cite{TCDCN}.
Denote by $I^{'}$ the horizontal flip of $I$.
Landmarks of $I^{'}$, denoted by $\left\{(x^{g'}_i,y^{g'}_i)\left| {_{i = 1}^{68}} \right.\right\}$, can be obtained by horizontal flip of $(x^g,y^g)$.
In order to align $I^{o'}$ to the pose of $I^o$, it is natural to require $(\Phi^x_{x^{g'}_i,y^{g'}_i}, \Phi^y_{x^{g'}_i,y^{g'}_i})$ be close to $(x^g_i, y^g_i)$, and we thus define the landmark loss as:
\begin{equation}\label{eq:landmark}
\ell_{lm} = \sum\limits_{i=1}^{68}(\Phi^x_{x^{g'}_i,y^{g'}_i}-x^{g}_i)^2+(\Phi^y_{x^{g'}_i,y^{g'}_i}-y^{g}_i)^2.
\end{equation}

Furthermore, TV regularization is deployed to constrain the spatial smoothness of flow field $\Phi$.
Given the 2D dense flow field $(\Phi^x,\Phi^y)$, the TV regularization is defined as:
\begin{equation}\label{eq:tv}
\ell_{TV}={\left\| {{\nabla_x}}\Phi^x \right\|^2}+{\left\| {{\nabla_y}}\Phi^x \right\|^2}+{\left\| {{\nabla_x}}\Phi^y \right\|^2}+{\left\| {{\nabla_y}}\Phi^y \right\|^2},
\end{equation}
where $\nabla_x$ and $\nabla_y$ denote the gradient operators along $x$ and $y$ coordinates, respectively.

\subsubsection{LightNet}

Generally, the left and right half-faces are lighting inconsistent,
%making
therefore, we cannot fill in missing pixels directly by $I^w$.
In order to compensate the illumination variation, we add the light adjustment module (LightNet) to make the completion result more harmonious.
LightNet takes $I^o$ and $I^{o'}$ as inputs, and adopts the same network architecture of FlowNet but it predicts the illumination ratio $R$ { as shown in the start of Sec.~\ref{sec::wholeNet}} as follows:
\begin{equation}
\label{eq:light}
R = \mathcal{F}_l(I^o,I^{o'}; \Theta_l),
\end{equation}
where $\Theta_l$ denotes the LightNet model parameters.

Given the illumination ratio $R$, warped image $I^w$, the inpainting result for missing pixels within only one of the half-face can be given by $M^{s1} \odot I^w \odot R$.
Taking the surrounding context into account, the completion result in the first stage can be obtained by:
\label{sec::IRW-Net}
\begin{equation}\label{eq:fusionimg}
\hat{I}^1 = M^{s1} \odot I^w \odot R + I^o \odot (1-M^{s1}).
\end{equation}
Here, $\odot$ represents the element-wise product operation.

We note that illumination reweighted warping cannot handle missing pixels happened on both of the half-faces, which will be addressed in the second stage.

For training LightNet, we introduce an illumination consistency loss.
Denote by $I^{w'}$ the warped version of the flip ground-truth $I^'$.
Then, the illumination reweighted $I^{w'}$ is required to approximate the original ground-truth $I$.
And we thus define the illumination consistency loss as:
\begin{equation}\label{eq:light2}
\mathcal{L}_l=\|{I^{w'} \odot R - I} \|^2.
\end{equation}

\subsection{Generative reconstruction}\label{sec::Rec-Net}

We further present a generative reconstruction subnet for the inpainting of missing pixels happened on both of the half-faces.
Let $M^{s2} = 1 - M - M^{s1}$.
%
%Then the missing pixels $(i, j)$ on both of the half-faces can be identified as $M^{s2}_{i,j} = 1$.
{When $M^{s2}_{i,j} = 1$, it indicates that pixel at location $(i, j)$ is missing. }
Thus, generative reconstruction subnet (RecNet) takes $\hat{I}^1$ (the completion result in the first stage) and $M^{s2}$ as input to generate the final completion result.
\begin{equation}\label{eq:sym}
\hat{I} = \mathcal{F}_r (\hat{I}^1,M^{s2}; \Theta_r),
\end{equation}
where $\Theta_r$ represents the RecNet model parameters.
For RecNet, we adopt the U-Net architecture~\cite{ronneberger2015u} which has the same structure with FlowNet.
Moreover, skip connections are included to concatenate each $l$-th layer to the $(L-l)$-th layer, where $L$ is the network depth.

The flow field $\Phi$ is further utilized to enforce the symmetry consistency on the completion results of missing pixels on both of the half-faces.
RecNet also takes the flip versions of $\hat{I}^1$ and $M^{s2}$ as input to generate $\hat{I}^{'}$.
We define $\Omega_l$ ($\Omega^'_l$) as the $(L-l)$-th layer of decoder feature map for $\hat{I}^1$ and $M^{s2}$ (their flip versions).
By downsampling $\Phi$ ($M^{s2}$) to $\Phi_{\downarrow}$ ($M^{s2}_{\downarrow}$) which has the same size with $\Omega_l$, the perceptual symmetry loss can then be defined as:
\begin{equation}\label{eq:symfeature}
\mathcal{L}_s \!\!=\!\! \frac{1}{C_l}\sum_{i,j}{( ({\Omega_l(i,j) \!-\! \Omega^'_l(\Phi^x_{\downarrow,i,j},\Phi^y_{\downarrow,i,j})}) \!\odot\! M^{s2}_{\downarrow}(i,j))^2},
\end{equation}
where $C_l$ denotes the channel number of the feature map $\Omega_l$.
In our implementation, we set $l=1$ with feature size $128 \times 128$.
Benefited from $\mathcal{L}_s$, we can maintain symmetric consistency even for filling in the missing pixels on both of the half-faces.

Reconstruction loss is introduced to require the final completion result $\hat{I}$ be close to the ground-truth $I$, which involves two terms.
The first one, $\ell_2$ loss, is defined as the squared Euclidean distance between $\hat{I}$ and $I$,
\begin{equation}\label{eq:l2}
\ell_2=\| {\hat{I}-I} \|^2.
\end{equation}
Inspired by~\cite{johnson2016perceptual}, the second term adopts the perceptual loss defined on pre-trained VGG-Face~\cite{parkhi2015deep}.
Denote by $\Psi$ the VGG-Face model, and $\Psi_k$ the $k$-th layer (\ie, $k=5$) of feature map.
The perceptual loss is then defined as,
\begin{equation}\label{eq:perceptual}
\ell_{perceptual} = \frac{1}{{{C_k}{H_k}{W_k}}}{\| {{\Psi _k}(\hat{I}) - {\Psi _k}(I)} \|^2},
\end{equation}
where the $C_k$, $H_k$ and $W_k$ denote the channel number, height and width of feature maps, respectively.
Then, the reconstruction loss is defined as,
\begin{equation}\label{recloss}
\mathcal{L}_r=\lambda_{r,2}\ell_2+\lambda_{r,p}\ell_{perceptual}.
\end{equation}
where $\lambda_{r,2}$ and $\lambda_{r,p}$ are the tradeoff parameters.

Finally, adversarial loss is deployed to generate photo-realistic completion result.
In~\cite{GFC2017CVPR,iizuka2017globally}, global and local discriminators are exploited, where local discriminator is defined on the inpainting result of a hole.
Considering that the hole may be irregular, it is inconvenient to define and learn local discriminator.
Instead, we apply local discriminators to four specific facial parts, \emph{i.e.}, left/right eye, nose and mouth.
Thus, local discriminators are consistent for any images with any missing masks, and facilitate the learning process of SymmFCNet.
For each part, we define its local adversarial loss as,
\begin{equation}\label{eq:part_discriminator}
\begin{split}
\ell_{a,p_i}=& \min_{\Theta} \max_{D_{p_i}} \mathbb{E}_{I_{p_i} \sim p_{data}({I_{p_i}})} [\log D_{p_i}({I_{p_i}})] + \\
& \mathbb{E}_{\hat{I}_{p_i} \sim p_{rec}({\hat{I}_{p_i}})} [\log ( 1-D_{p_i}(\hat{I}_{p_i}) )],
\end{split}
\end{equation}
where $p_{data}(I_{p_i})$ and $p_{rec}(\hat{I}_{p_i})$ stands for the distributions of the $i$-th part from $I$ and $\hat{I}$, respectively. $D_{p_i}$ denotes the $i$-th part discriminator.
To sum up, the overall adversarial loss is defined as,
\begin{equation}\label{eq:adv}
\mathcal{L}_a=\lambda_{a,g}\ell_{a,g} + \sum_i \lambda_{a,p_i} \ell_{a,p_i},
\end{equation}
where $\ell_{a,g}$ represents the global adversarial loss~\cite{goodfellow2014generative} working on the whole image rather than parts, $\lambda_{a,g}$ and $\lambda_{a,p_i}$ are the tradeoff parameters for the global and local adversarial losses, respectively.
Here, left eye, right eye, nose, and mouth denote the first, second, third and fourth parts, respectively.
For each part cropped from face images, we employ bi-linear interpolation to resize it to $128\times128$.

\subsection{Learning Objective}

Taking all the losses on FlowNet, LightNet and RecNet into account, the overall objective of SymmFCNet can be defined as,
\begin{equation}\label{eq:all}
\mathcal{L} = \mathcal{L}_r + \mathcal{L}_a + \lambda_s \mathcal{L}_s + \lambda_l \mathcal{L}_l + \lambda_{lm} \mathcal{L}_{lm} + \lambda_{TV} \mathcal{L}_{TV},
\end{equation}
where $\lambda_s$, $\lambda_l$, $\lambda_{lm}$ and $\lambda_{TV}$ are the tradeoff parameters for symmetry consistency loss, illumination consistency loss, landmark loss and TV regularization, respectively.
Note that our SymmFCNet is constructed by stacking generative reconstruction subnet upon illumination reweighted warping subnet and can be trained in an end-to-end manner.
Thus, FlowNet and LightNet can also be learned by minimizing $\mathcal{L}_r$, $\mathcal{L}_a$ and $\mathcal{L}_s$ even they are defined on RecNet.

\section{Experiments}\label{result}

In this section, experiments are conducted to assess our SymmFCNet and compare it with the state-of-the-art image inpainting and face completion methods~\cite{GFC2017CVPR,iizuka2017globally,Yu2018CVPR,liu2018image}.
For comprehensive evaluation, quantitative and qualitative results as well as user study are reported.
In addition, we test the completion performance on both images with synthetic missing pixels and images with real occlusion.
%
%More results are included in the suppl.
%
%The pre-trained models and source code will be publicly available.% after the acceptance of the paper.
Testing code is available at: \url{https://github.com/csxmli2016/SymmFCNet}.

\subsection{Dataset and Setting}
The VGGFace2 dataset~\cite{Vggface2} is used to train our SymmFCNet.
The dataset contains 9,131 identities and each has an average of 362 images, from which we manually select 19,000 images to constitute our training set by excluding images with low quality and large occlusions.
A validation set is also built by selecting another 400 images from VGGFace2 for guiding the settings of model and learning parameters.
We adopt two test sets to assess SymmFCNet.
The first one involves 1,200 images from VGGFace2, and the other includes 1,200 images from WebFace~\cite{Webface} to verify generalization performance across datasets.
The identities of face images from training, validation and test sets are non-overlapped.
Using bounding box detected by MTCNN~\cite{MTCNN}, each face image is cropped and resized to $256\times256$.

%\subsection{Implementation Details}

The model parameters for SymmFCNet are set as follows: $\lambda_{r,2}=300$, $\lambda_{r,p}=0.01$, $\lambda_{a,g}=100$, $\lambda_{a,p_1} = \lambda_{a,p_2} =100$, $\lambda_{a,p_3} = \lambda_{a,p_4} = 80$, $\lambda_{s} = 50$, $\lambda_{lm} = 10$, $\lambda_{TV}=1$, $\lambda_{l}=100$.
The
%missing
pixel-missing masks are generated by randomly selecting the location and mask size.
Data augmentation such as flipping and random cropping are also adopted.
The training of SymmFCNet includes three stages.
(i) We first pre-train illumination reweighted warping subnet for 10 epochs. (ii) Fixed FlowNet and LightNet, we pre-train RecNet for 20 epochs. (iii) Finally, the full SymmFCNet is end-to-end trained by minimizing the learning objective $\mathcal{L}$.
To train SymmFCNet, we use the ADAM algorithm~\cite{kingma2014adam} with the learning rate of $2\times10^{-4}$, $2\times10^{-5}$, $2\times10^{-6}$ and $\beta_1=0.5$,
where a smaller learning rate is adopted until the reconstruction loss $\mathcal{L}_r$ on validation set becomes non-decreasing.
To improve the perception quality, the tradeoff parameters of adversarial losses are gradually increased according to $\mathcal{L}_r$ on validation set.
The batch size is 1 and the training is stopped after 200 epochs.

\subsection{Results on Images with Synthetic Missing Pixels}
Quantitative and qualitative results are reported on our SymmFCNet and four state-of-the-art methods \cite{GFC2017CVPR,iizuka2017globally,Yu2018CVPR,liu2018image}.
Among them, Li \emph{et al.}~\cite{GFC2017CVPR} and Iizuka \emph{et al.}~\cite{iizuka2017globally} can only handle $128\times128$ images, and we use bicubic interpolation to upsample the output to the size of $256\times256$.
For Iizuka \emph{et al.}~\cite{iizuka2017globally}, we exploit the alignment tool suggested by the authors to pre-process the input image.
For Liu \etal~\cite{liu2018image}, it upsamples the input to $512\times512$ and we downsample the output to $256\times256$.
Online manual specification of missing masks is required to obtain the results by Liu \etal~\cite{liu2018image}, and we thus do not report its quantitative metrics (\emph{e.g.}, PSNR) because it is exhausted to manually edit the masks for thousands of images.

\subsubsection{Quantitative Results}\label{quantitative}

\begin{table}[t]
%\small
%\footnotesize
\scriptsize
%\tiny
%\small
\setlength{\abovecaptionskip}{4pt}
\setlength{\belowcaptionskip}{-6pt}
\begin{center}
\setlength{\tabcolsep}{0.12mm}%1mm
{
\begin{tabular}{|c| c| c c c c| c c c c|}
\hline%SSIM & Ver(\%)
\multicolumn{2}{|c|}{\multirow{2}{*}{\makecell[c]{Methods}}} & \multicolumn{4}{c|}{VggFace2~\cite{Vggface2}} & \multicolumn{4}{c|}{WebFace~\cite{Webface}}\\
\cline{3-10}
\multicolumn{2}{|c|}{}& PSNR$\uparrow$ & SSIM$\uparrow$ & LPIPS$\downarrow$ & Dis.$\downarrow$ & PSNR$\uparrow$ & SSIM$\uparrow$ & LPIPS$\downarrow$ & Dis.$\downarrow$ \\
\hline\hline
\multirow{3}{*}{\shortstack{State-of-\\ the-arts}}&Iizuka \emph{et al.}~\cite{iizuka2017globally}& 18.62 & .688 & .513 & 1.325 &  19.04 & .683 & .504 & 1.462  \\
& Li \emph{et al.}~\cite{GFC2017CVPR}& 25.05 & .932 & .397 & 0.932 & 25.65 & .959 & .371 & 1.116  \\
& Yu \emph{et al.}~\cite{Yu2018CVPR}& 25.53 & .963 & .292  & 0.788 & 25.96 & .965 & .270 &  0.965 \\
\hline
%Ours (\emph{V1})& 25.09 & .955 & .274 & 0.938 & 25.36 & .957 & .291 & 1.095 \\
\multirow{4}{*}{\shortstack{Ablation\\ study}}& Plain RecNet& 24.99 & .957 & .279 & 0.864 & 25.81 & .959 & .310 & 1.035 \\
& SymmFcNet (\emph{-GL0}) & 25.54 & .959 & .260 & 0.830 & 25.94 & .963 & .294 & 0.987 \\
& SymmFcNet (\emph{-L}) & 26.43 & .967 & .226 & 0.622  & 26.43 & .968 & .258 & 0.852 \\
&SymmFCNet (\emph{-S})& 26.33 & .962 & .232 & 0.714 & 26.17 & .961 & .266 & 0.945 \\
\hline
\multicolumn{2}{|c|}{\textbf{SymmFCNet (\emph{Full})}} & \textbf{27.81} & \textbf{.970} & \textbf{.219} & \textbf{0.617} & \textbf{27.22} & \textbf{.969} & \textbf{.252}  & \textbf{0.849}\\
\hline
\end{tabular}}
\end{center}
\caption{Quantitative results. Here, $\uparrow$ ($\downarrow$) indicates higher (lower) is better.}
\label{tab:quan}
\end{table}

Table~\ref{tab:quan} lists the PSNR, SSIM, identity distance (Dis.) by OpenFace toolbox~\cite{amos2016openface}, and perceptual similarity (LPIPS)~\cite{zhang2018perceptual} on the the two test sets (\emph{i.e.}, VGGFace and WebFace).
In comparison with the competing methods, notable PSNR gain (\emph{i.e.}, $>$1 dB) is achieved by our SymmFCNet.
In terms of SSIM, our SymmFCNet also performs favorably.
LPIPS~\cite{zhang2018perceptual} is a recently proposed perceptual similarity which is more consistent with human perception.
Again our SymmFCNet achieves the best LPIPS performance in comparison to the competing methods.
In addition, identity distance measures whether the result and ground-truth have the same identity, and thus can be used to assess the coherence of the completion result with surrounding context.
From Table~\ref{tab:quan}, it can be seen that SymmFCNet exhibits better identity-preserving ability than the competing methods.

\subsubsection{Qualitative Results}

\begin{figure*}[!t]
%\vspace{-5pt}
\centering
\subfigure[Input]{
  \begin{minipage}[b]{.292\columnwidth}
    \includegraphics[width=0.99\textwidth]{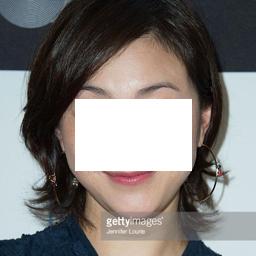}\\
    \includegraphics[width=0.99\textwidth]{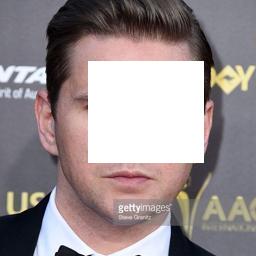}
    \includegraphics[width=0.99\textwidth]{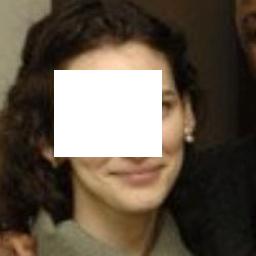}\\
    \includegraphics[width=0.99\textwidth]{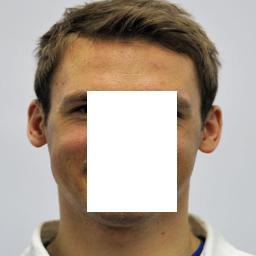}
  \end{minipage}
}
\hspace{-2.1ex}
\subfigure[Iizuka \emph{et al.}~\cite{iizuka2017globally}]{
  \begin{minipage}[b]{.292\columnwidth}
    \includegraphics[width=0.99\textwidth]{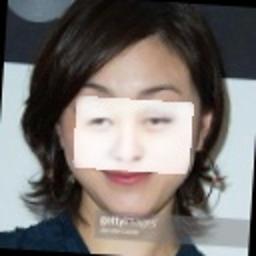}\\
    \includegraphics[width=0.99\textwidth]{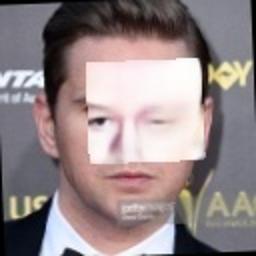}
    \includegraphics[width=0.99\textwidth]{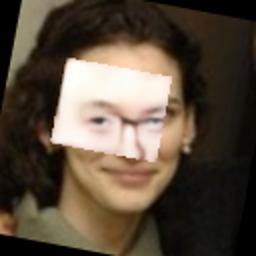}\\
    \includegraphics[width=0.99\textwidth]{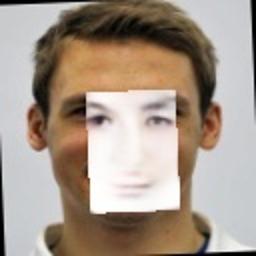}
  \end{minipage}
}
\hspace{-2.1ex}
\subfigure[Li \emph{et al.}~\cite{GFC2017CVPR}]{
  \begin{minipage}[b]{.292\columnwidth}
    \includegraphics[width=0.99\textwidth]{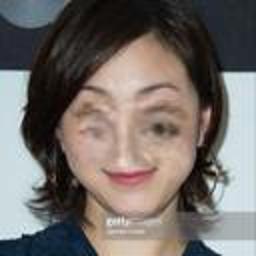}\\
    \includegraphics[width=0.99\textwidth]{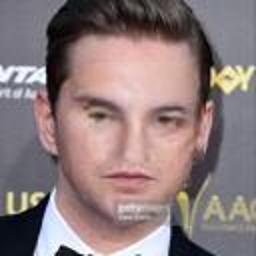}
    \includegraphics[width=0.99\textwidth]{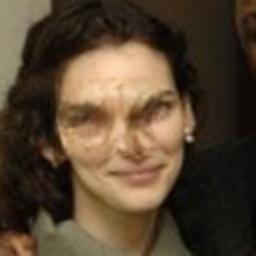}\\
    \includegraphics[width=0.99\textwidth]{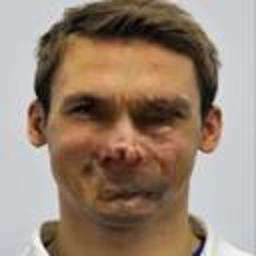}
  \end{minipage}
}
\hspace{-2.1ex}
\subfigure[Yu \emph{et al.}~\cite{Yu2018CVPR}]{
  \begin{minipage}[b]{.292\columnwidth}
    \includegraphics[width=0.99\textwidth]{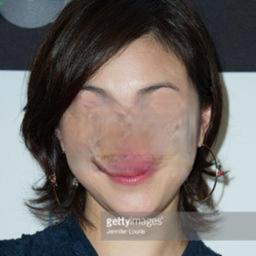}\\
    \includegraphics[width=0.99\textwidth]{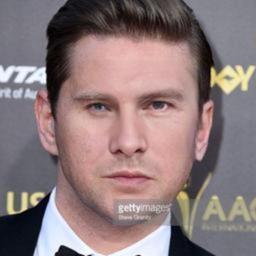}
    \includegraphics[width=0.99\textwidth]{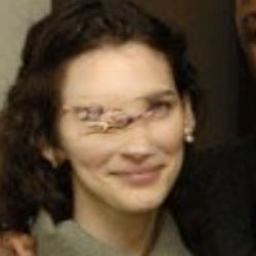}\\
    \includegraphics[width=0.99\textwidth]{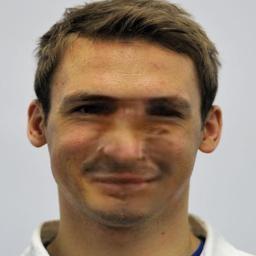}
  \end{minipage}
}
\hspace{-2.1ex}
\subfigure[Liu \emph{et al.}~\cite{liu2018image}]{
  \begin{minipage}[b]{.292\columnwidth}
    \includegraphics[width=0.99\textwidth]{./eps/RegularMask/PConv/n0076530003_L_92_103_21_R_162_98_24_N_0_0_0_NewN_129_153_29_M_133_167_29_NewBB_0_0_0_0.jpg}\\
    \includegraphics[width=0.99\textwidth]{./eps/RegularMask/PConv/n0003520001_L_99_103_25_R_166_107_23_N_0_0_0_NewN_134_144_26_M_130_177_24_NewBB_0_0_0_0.jpg}
    \includegraphics[width=0.99\textwidth]{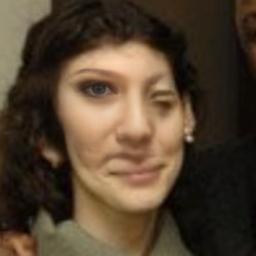}\\
    \includegraphics[width=0.99\textwidth]{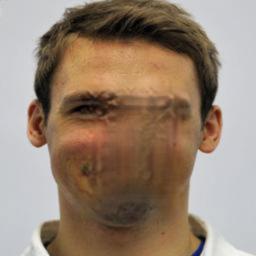}
  \end{minipage}
}
\hspace{-2.1ex}
\subfigure[Ours]{
  \begin{minipage}[b]{.292\columnwidth}
    \includegraphics[width=0.99\textwidth]{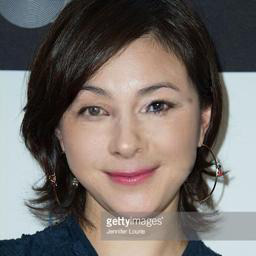}\\
    \includegraphics[width=0.99\textwidth]{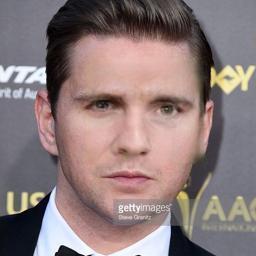}
    \includegraphics[width=0.99\textwidth]{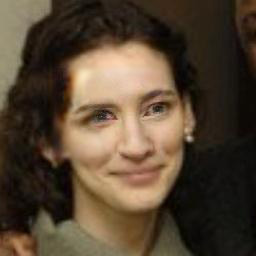}\\
    \includegraphics[width=0.99\textwidth]{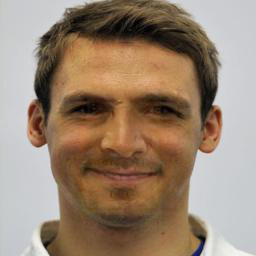}
  \end{minipage}
}
\hspace{-2.1ex}
\subfigure[Ground-truth]{
  \begin{minipage}[b]{.292\columnwidth}
    \includegraphics[width=0.99\textwidth]{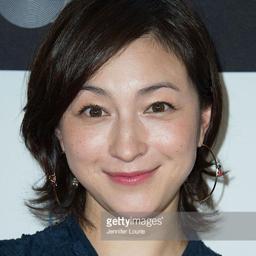}\\
    \includegraphics[width=0.99\textwidth]{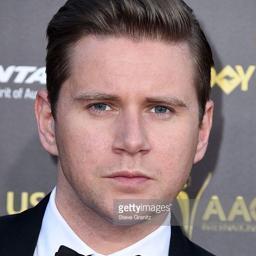}
    \includegraphics[width=0.99\textwidth]{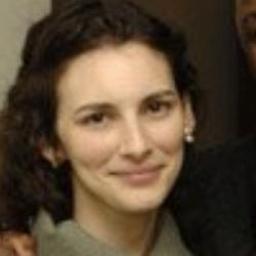}\\
    \includegraphics[width=0.99\textwidth]{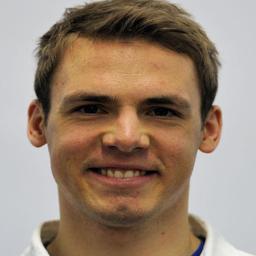}
  \end{minipage}
}
  \caption{Completion results on regular holes.}
  \label{fig:symmetryR}
\end{figure*}

\begin{figure*}[t]
\centering
\subfigure[Input]{
  \begin{minipage}[b]{.292\columnwidth}
    \includegraphics[width=0.99\textwidth]{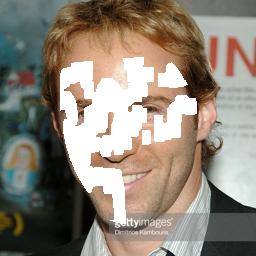}\\
    \includegraphics[width=0.99\textwidth]{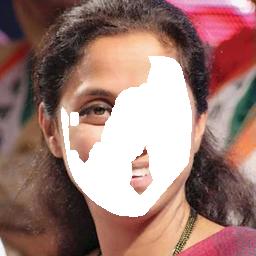}
    \includegraphics[width=0.99\textwidth]{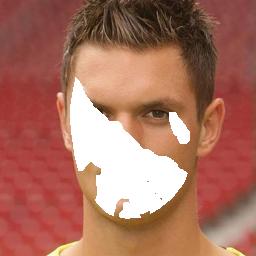}\\
    \includegraphics[width=0.99\textwidth]{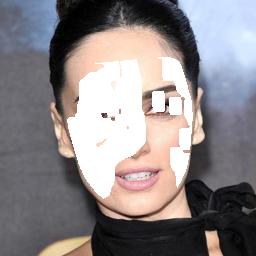}
  \end{minipage}
}
\hspace{-2.1ex}
\subfigure[Iizuka \emph{et al.}~\cite{iizuka2017globally}]{
  \begin{minipage}[b]{.292\columnwidth}
    \includegraphics[width=0.99\textwidth]{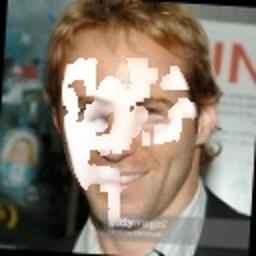}\\
    \includegraphics[width=0.99\textwidth]{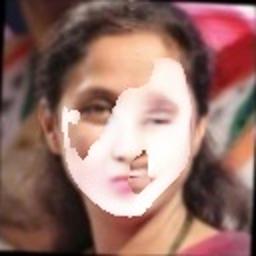}
    \includegraphics[width=0.99\textwidth]{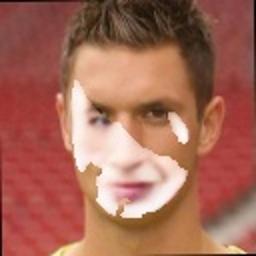}\\
    \includegraphics[width=0.99\textwidth]{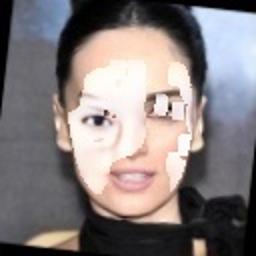}

  \end{minipage}
}
\hspace{-2.1ex}
\subfigure[Li \emph{et al.}~\cite{GFC2017CVPR}]{
  \begin{minipage}[b]{.292\columnwidth}
    \includegraphics[width=0.99\textwidth]{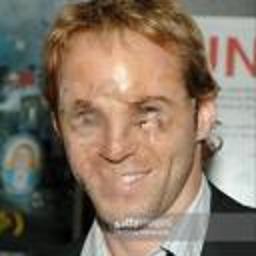}\\
    \includegraphics[width=0.99\textwidth]{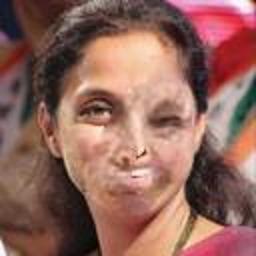}
    \includegraphics[width=0.99\textwidth]{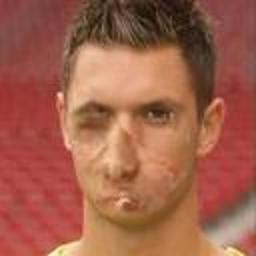}\\
    \includegraphics[width=0.99\textwidth]{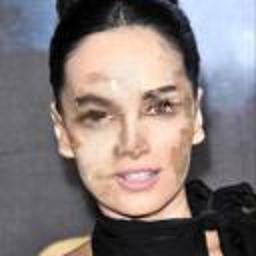}
  \end{minipage}
}
\hspace{-2.1ex}
\subfigure[Yu \emph{et al.}~\cite{Yu2018CVPR}]{
  \begin{minipage}[b]{.292\columnwidth}
    \includegraphics[width=0.99\textwidth]{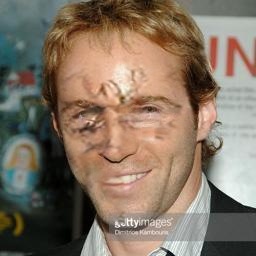}\\
    \includegraphics[width=0.99\textwidth]{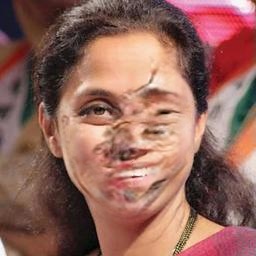}
    \includegraphics[width=0.99\textwidth]{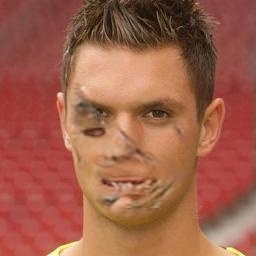}\\
    \includegraphics[width=0.99\textwidth]{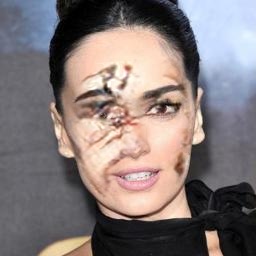}
  \end{minipage}
}
\hspace{-2.1ex}
\subfigure[Liu \emph{et al.}~\cite{liu2018image}]{
  \begin{minipage}[b]{.292\columnwidth}
    \includegraphics[width=0.99\textwidth]{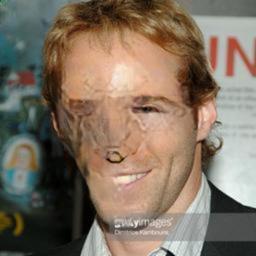}\\
    \includegraphics[width=0.99\textwidth]{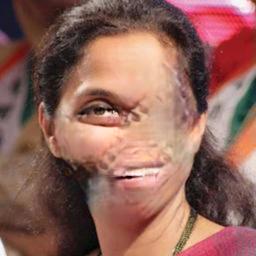}
    \includegraphics[width=0.99\textwidth]{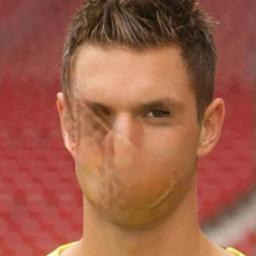}\\
    \includegraphics[width=0.99\textwidth]{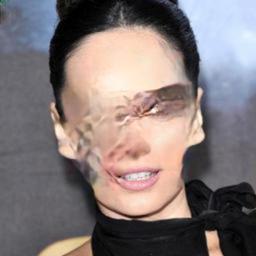}

  \end{minipage}
}
\hspace{-2.1ex}
\subfigure[Ours]{
  \begin{minipage}[b]{.292\columnwidth}
    \includegraphics[width=0.99\textwidth]{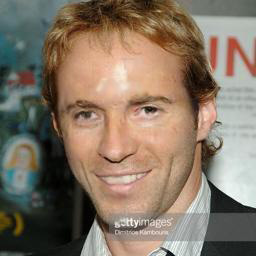}\\
    \includegraphics[width=0.99\textwidth]{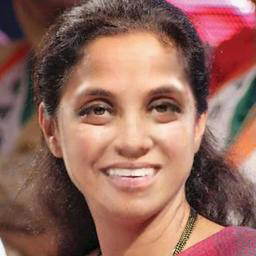}
    \includegraphics[width=0.99\textwidth]{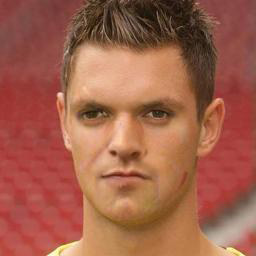}\\
    \includegraphics[width=0.99\textwidth]{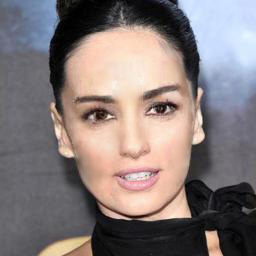}
  \end{minipage}
  }
\hspace{-2.1ex}
\subfigure[Ground-truth]{
  \begin{minipage}[b]{.292\columnwidth}
    \includegraphics[width=0.99\textwidth]{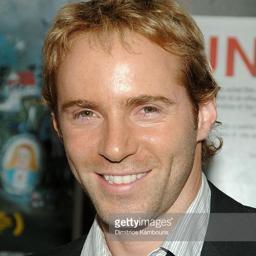}\\
    \includegraphics[width=0.99\textwidth]{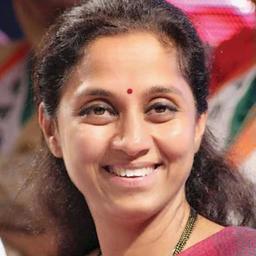}
    \includegraphics[width=0.99\textwidth]{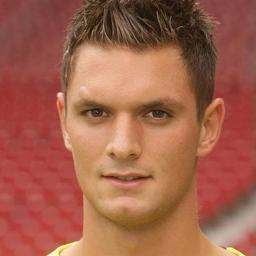}\\
    \includegraphics[width=0.99\textwidth]{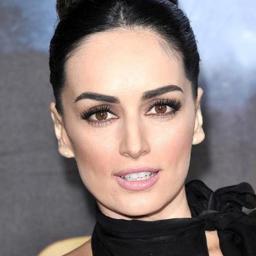}
  \end{minipage}
}
  \caption{Completion results on irregular holes.}
  \label{fig:irregular}
\end{figure*}

The solutions for face completion are neither unique  nor required to exactly approximate the ground-truth.
Thus, qualitative comparison is conducted to show the effectivenss of our methods.
%present an intuitive illustration of the results.
%
Figs.~\ref{fig:symmetryR} and~\ref{fig:irregular} show the completion results on rectangular and irregular holes, respectively.
The results by Liu \etal~\cite{liu2018image} are also included for comparison.
{Benefited from the joint effectiveness of illumination reweighted warping and perceptual symmetric loss, our SymmFCNet can achieve very promising inpainting results which preserve visually pleasing symmetry consistent details for missing pixels within only one half-faces and both half-faces.}
In comparison, the methods~\cite{GFC2017CVPR,Yu2018CVPR,iizuka2017globally} fail to recover rich details and even semantic facial structures, while Liu \etal~\cite{liu2018image} is still limited in maintaining global symmetry consistency and sometimes fails in generating plausible results with large occlusions.

\subsubsection{{User Study}}
\begin{table}[t]
\small
%\footnotesize
%\scriptsize
\setlength{\abovecaptionskip}{4pt}
\setlength{\belowcaptionskip}{-8pt}
\begin{center}
\setlength{\tabcolsep}{1.1mm}%1mm
{
\begin{tabular}{|c| c c c|}
\hline%SSIM & Ver(\%)
Methods & Regular Mask & Irregular Mask & Real Image \\
\hline\hline
Iizuka \emph{et al.}~\cite{iizuka2017globally} & 2.04\% & 3.52\% & 2.56\% \\
Li \emph{et al.}~\cite{GFC2017CVPR} & 0.24\% & 0.16\% & 0.48\% \\
Yu \emph{et al.}~\cite{Yu2018CVPR} & 0.40\% & 0.64\% & 2.40\% \\
Liu \emph{et al.}~\cite{liu2018image} & 16.80\% & 1.68\% & 15.04\% \\
\hline
\textbf{SymmFCNet} & \textbf{80.52\%} & \textbf{94.00\%} & \textbf{79.52 \%}\\
\hline
\end{tabular}}
\end{center}
\caption{{Voting results for three types of missing pixels}.}
\label{tab:us}
\end{table}
{User study is conducted on a crowdsourcing platform for three types of missing pixels, \emph{i.e.}, regular mask, irregular mask and real occlusions, which contain 50, 25 and 25 images, respectively.
For each image, we display the results by our SymmFCNet and the methods~\cite{GFC2017CVPR,Yu2018CVPR,iizuka2017globally,liu2018image} in random order to 50 workers who are required to choose the one with the best global consistency and perception quality.
We use the percent of the votes of one particular algorithm against all votes to evaluate the performance of the algorithm in Table.~\ref{tab:us}.
The result by SymmFCNet has $84.68\%$ probability on average to be selected as the best one.}

\subsubsection{Running Time}

All the experiments are conducted on a computer equipped with Intel Xeon E3 CPU and NVIDIA GeForce GTX 1080Ti GPU. And the model is trained and tested with Torch. Our SymmFCNet takes 36.29 $ms$ on average for completing a $256\times256$ image.

\subsection{Results on Images with Real Occlusions}

\begin{figure*}[t]
\centering
\subfigure[Real Image]{
  \begin{minipage}[b]{.34\columnwidth}
    \includegraphics[width=0.99\textwidth]{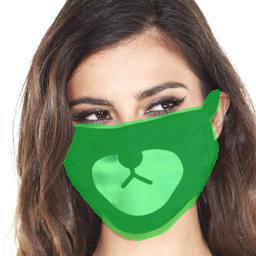}\\
    \includegraphics[width=0.99\textwidth]{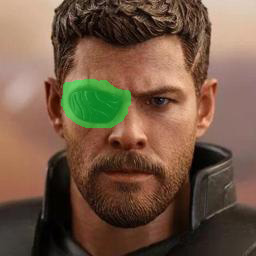}\\
    \includegraphics[width=0.99\textwidth]{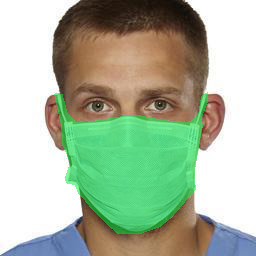}\\
    \includegraphics[width=0.99\textwidth]{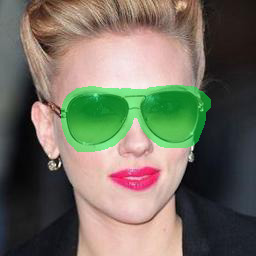}

  \end{minipage}
}
\hspace{-2.1ex}
\subfigure[Iizuka \emph{et al.}~\cite{iizuka2017globally}]{
  \begin{minipage}[b]{.34\columnwidth}
    \includegraphics[width=0.99\textwidth]{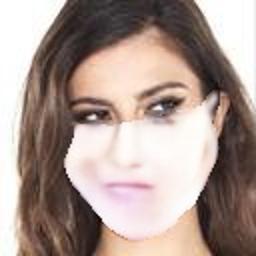}\\
    \includegraphics[width=0.99\textwidth]{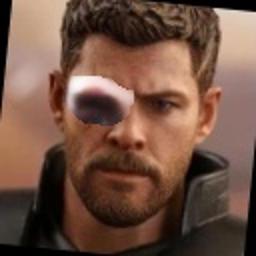}\\
    \includegraphics[width=0.99\textwidth]{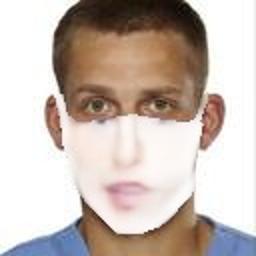}\\
    \includegraphics[width=0.99\textwidth]{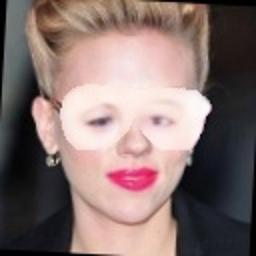}

  \end{minipage}
}
\hspace{-2.1ex}
\subfigure[Li \emph{et al.}~\cite{GFC2017CVPR}]{
  \begin{minipage}[b]{.34\columnwidth}
    \includegraphics[width=0.99\textwidth]{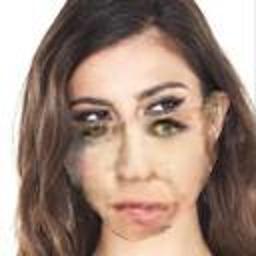}\\
    \includegraphics[width=0.99\textwidth]{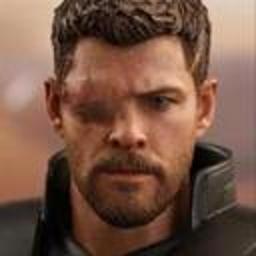}\\
    \includegraphics[width=0.99\textwidth]{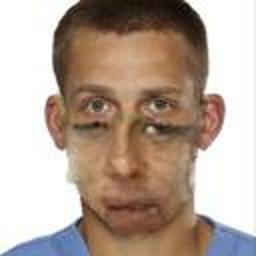}\\
    \includegraphics[width=0.99\textwidth]{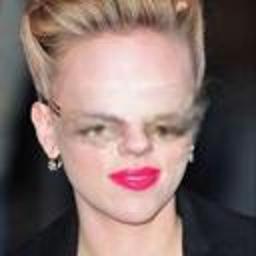}

  \end{minipage}
}
\hspace{-2.1ex}
\subfigure[Yu \emph{et al.}~\cite{Yu2018CVPR}]{
  \begin{minipage}[b]{.34\columnwidth}
    \includegraphics[width=0.99\textwidth]{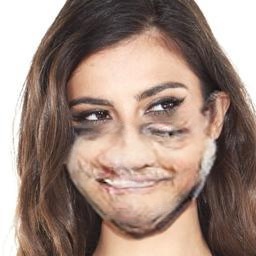}\\
    \includegraphics[width=0.99\textwidth]{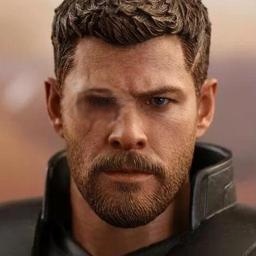}\\
    \includegraphics[width=0.99\textwidth]{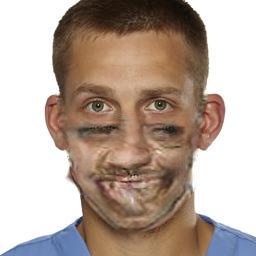}\\
    \includegraphics[width=0.99\textwidth]{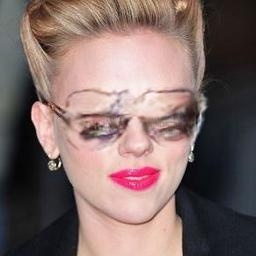}

  \end{minipage}
}
\hspace{-2.1ex}
\subfigure[Liu \emph{et al.}~\cite{liu2018image}]{
  \begin{minipage}[b]{.34\columnwidth}
    \includegraphics[width=0.99\textwidth]{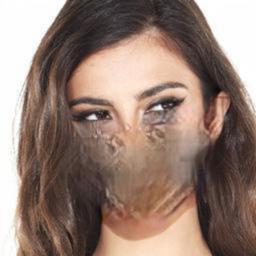}\\
    \includegraphics[width=0.99\textwidth]{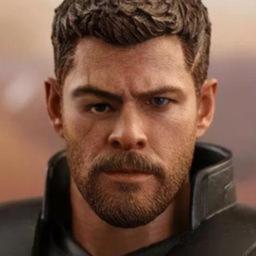}\\
    \includegraphics[width=0.99\textwidth]{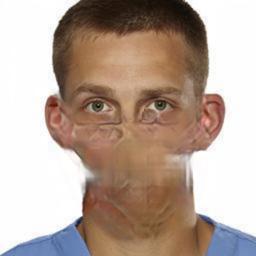}\\
    \includegraphics[width=0.99\textwidth]{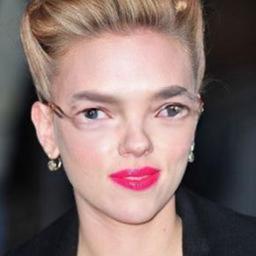}

  \end{minipage}
}
\hspace{-2.1ex}
\subfigure[Ours]{
  \begin{minipage}[b]{.34\columnwidth}
    \includegraphics[width=0.99\textwidth]{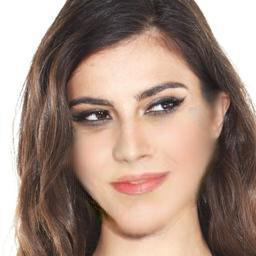}\\
    \includegraphics[width=0.99\textwidth]{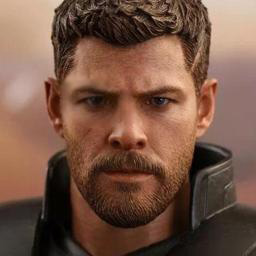}\\
    \includegraphics[width=0.99\textwidth]{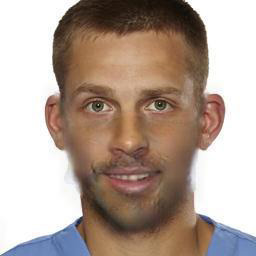}\\
    \includegraphics[width=0.99\textwidth]{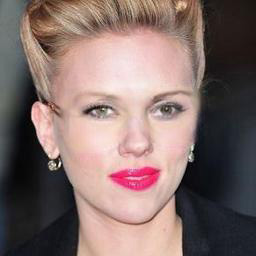}

  \end{minipage}
}
  \caption{Completion results on images with real occlusion.}
  \label{fig:RealImg}
\end{figure*}
By manually specifying the missing masks, Fig.~\ref{fig:RealImg} shows the completion results on two face images with real occlusions.
{For the first image, even the occlusion is large and nearly symmetric, SymmFCNet still performs favorably, validating the effectiveness of perceptual symmetry loss.
As for the second image, the occlusion is mainly in one half-face, and the result by SymmFCNet is globally more symmetry consistent in comparison to Liu \etal~\cite{liu2018image}.}
\begin{figure}[t]
\setlength{\abovecaptionskip}{8pt}
\setlength{\belowcaptionskip}{-10pt}
\centering
\subfigure[]{
  \begin{minipage}[b]{.195\columnwidth}
    \includegraphics[width=0.99\textwidth]{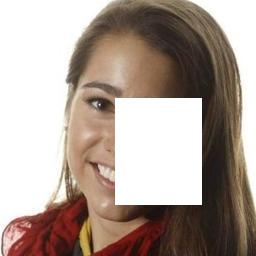}\\
    \includegraphics[width=0.99\textwidth]{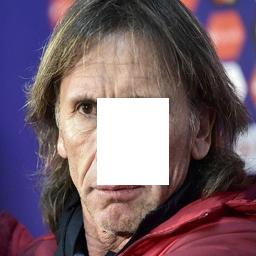}\\
    \includegraphics[width=0.99\textwidth]{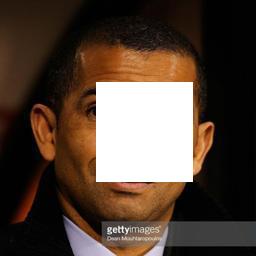}
  \end{minipage}
}
\hspace{-2.1ex}
\subfigure[]{
  \begin{minipage}[b]{.195\columnwidth}
    \includegraphics[width=0.99\textwidth]{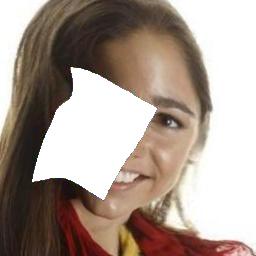}\\
    \includegraphics[width=0.99\textwidth]{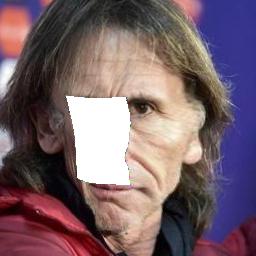}\\
    \includegraphics[width=0.99\textwidth]{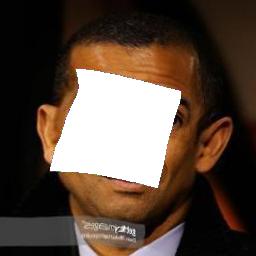}
  \end{minipage}
}
\hspace{-2.1ex}
\subfigure[]{
  \begin{minipage}[b]{.195\columnwidth}
    \includegraphics[width=0.99\textwidth]{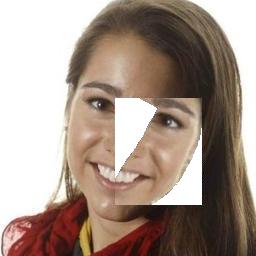}\\
    \includegraphics[width=0.99\textwidth]{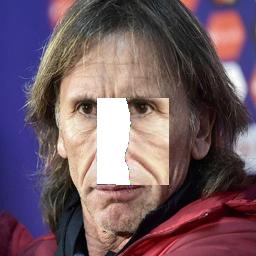}\\
    \includegraphics[width=0.99\textwidth]{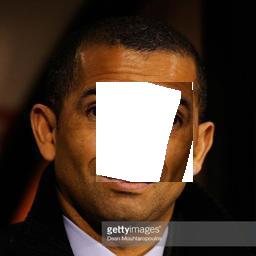}
  \end{minipage}
}
\hspace{-2.1ex}
\subfigure[]{
  \begin{minipage}[b]{.195\columnwidth}
    \includegraphics[width=0.99\textwidth]{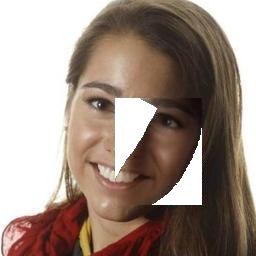}\\
    \includegraphics[width=0.99\textwidth]{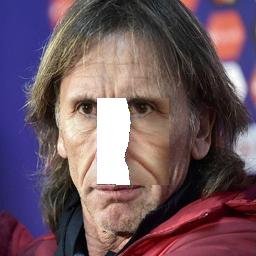}\\
    \includegraphics[width=0.99\textwidth]{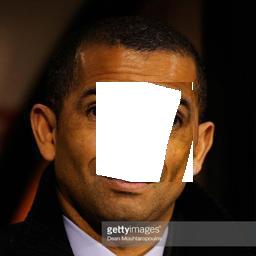}
  \end{minipage}
}
\hspace{-2.1ex}
\subfigure[]{
  \begin{minipage}[b]{.195\columnwidth}
    \includegraphics[width=0.99\textwidth]{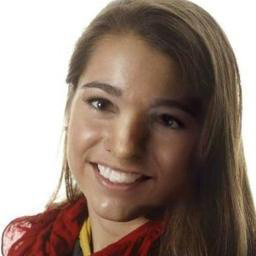}\\
    \includegraphics[width=0.99\textwidth]{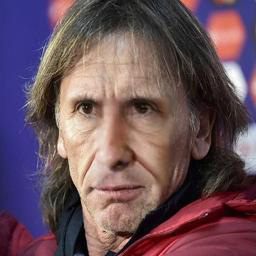}\\
    \includegraphics[width=0.99\textwidth]{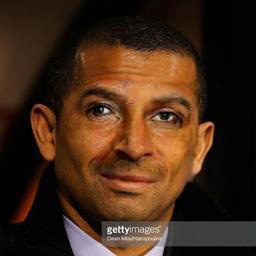}
  \end{minipage}
}
  \caption{Intermediate results of SymmFCNet, (a) occluded face $I^o$, (b) warped image $I^w$ by FlowNet, (c) completion result in the first stage without illumination correction, (d) completion result in the first stage after illumination correction, (e) final completion result $\hat{I}$ from RecNet.}
  \label{fig:stage1}
\end{figure}

\subsection{Ablation Study}\label{ablative}

\begin{figure*}[t]

\centering
\subfigure[]{
  \begin{minipage}[b]{.292\columnwidth}
    \includegraphics[width=0.99\textwidth]{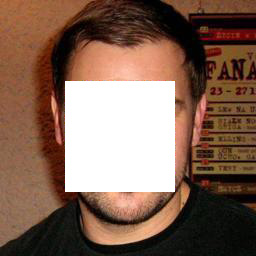}\\
    \includegraphics[width=0.99\textwidth]{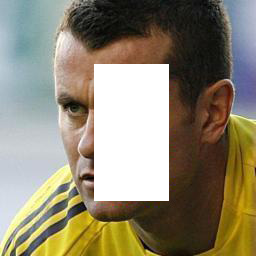}\\
    \includegraphics[width=0.99\textwidth]{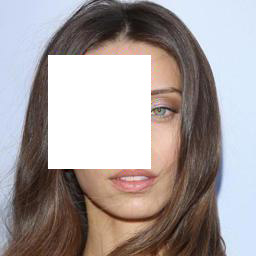}\\
    \includegraphics[width=0.99\textwidth]{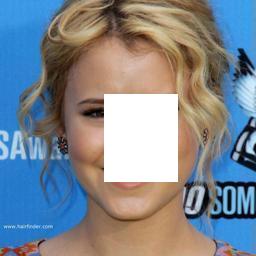}
  \end{minipage}
}
\hspace{-2.1ex}
\subfigure[]{
  \begin{minipage}[b]{.292\columnwidth}
    \includegraphics[width=0.99\textwidth]{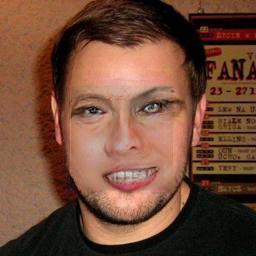}\\
    \includegraphics[width=0.99\textwidth]{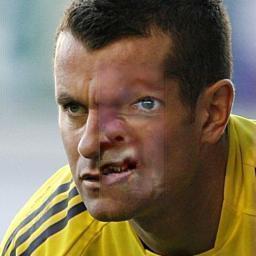}\\
    \includegraphics[width=0.99\textwidth]{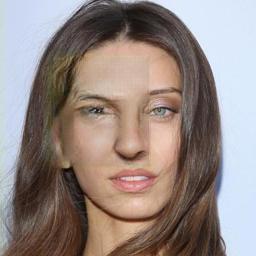}\\
    \includegraphics[width=0.99\textwidth]{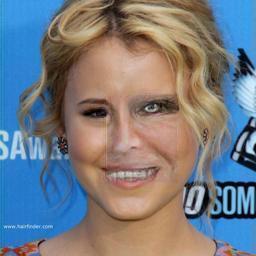}
  \end{minipage}
}
\hspace{-2.1ex}
\subfigure[]{
  \begin{minipage}[b]{.292\columnwidth}
    \includegraphics[width=0.99\textwidth]{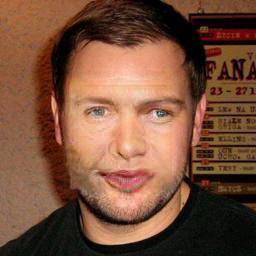}\\
    \includegraphics[width=0.99\textwidth]{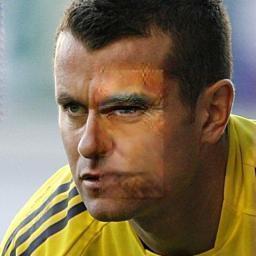}\\
    \includegraphics[width=0.99\textwidth]{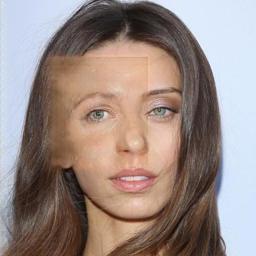}\\
    \includegraphics[width=0.99\textwidth]{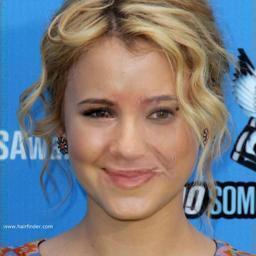}
  \end{minipage}
}
\hspace{-2.1ex}
\subfigure[]{
  \begin{minipage}[b]{.292\columnwidth}
    \includegraphics[width=0.99\textwidth]{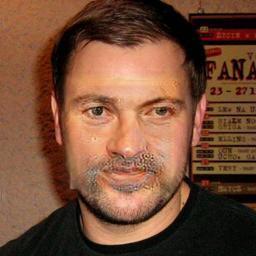}\\
    \includegraphics[width=0.99\textwidth]{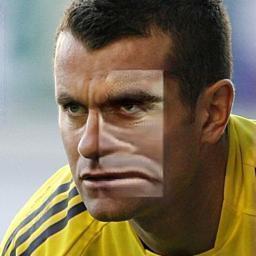}\\
    \includegraphics[width=0.99\textwidth]{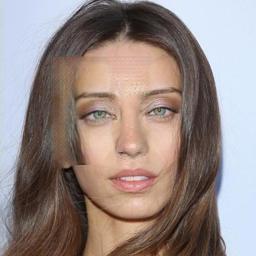}\\
    \includegraphics[width=0.99\textwidth]{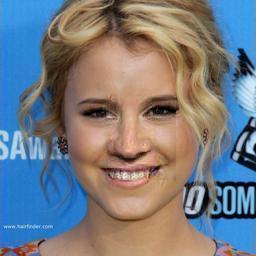}
  \end{minipage}
}
\hspace{-2.1ex}
\subfigure[]{
  \begin{minipage}[b]{.292\columnwidth}
    \includegraphics[width=0.99\textwidth]{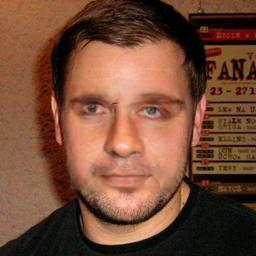}\\
    \includegraphics[width=0.99\textwidth]{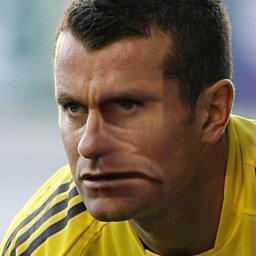}\\
    \includegraphics[width=0.99\textwidth]{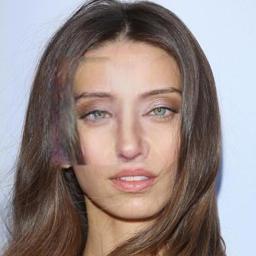}\\
    \includegraphics[width=0.99\textwidth]{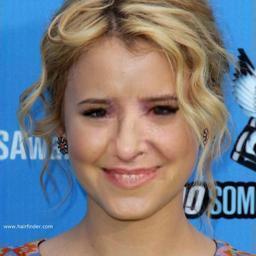}
  \end{minipage}
}
\hspace{-2.1ex}
\subfigure[]{
  \begin{minipage}[b]{.292\columnwidth}
    \includegraphics[width=0.99\textwidth]{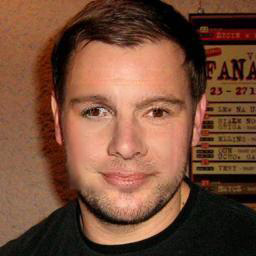}\\
    \includegraphics[width=0.99\textwidth]{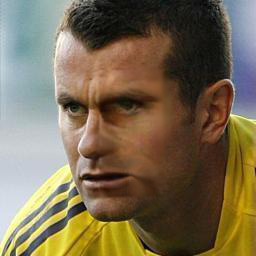}\\
    \includegraphics[width=0.99\textwidth]{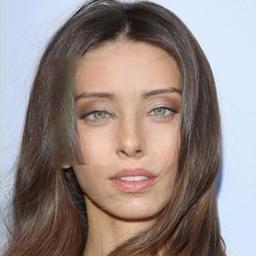}\\
    \includegraphics[width=0.99\textwidth]{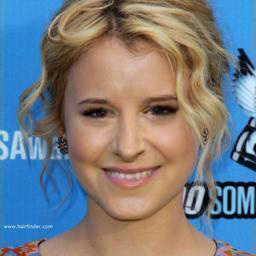}
  \end{minipage}
}
\hspace{-2.1ex}
\subfigure[]{
  \begin{minipage}[b]{.292\columnwidth}
    \includegraphics[width=0.99\textwidth]{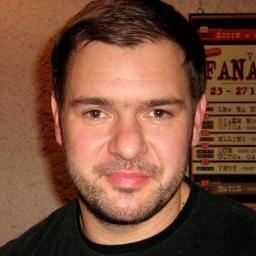}\\
    \includegraphics[width=0.99\textwidth]{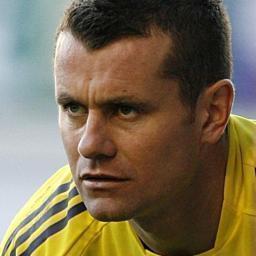}\\
    \includegraphics[width=0.99\textwidth]{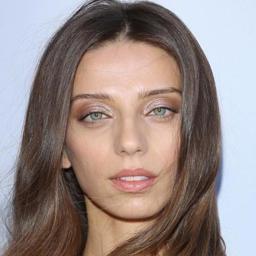}\\
    \includegraphics[width=0.99\textwidth]{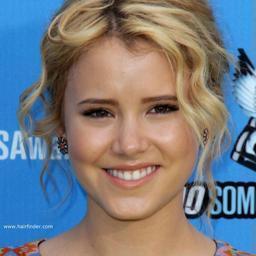}
  \end{minipage}
}
  \caption{Results of our SymmFCNet variants.
  From left to right: (a)~input, (b)~RecNet, (c)~SymmFCNet~(\emph{-GL0}), (d)~SymmFCNet~(\emph{-L}), (e)~SymmFCNet~(\emph{-S}), (f)~SymmFCNet~(\emph{Full}), (g)~ground-truth. Best viewed by zooming in the screen.}
  \label{fig:ablative}
\end{figure*}

Two groups of experiments are conducted to assess the contributions of main components in our SymmFCNet.
In the first group of experiments, Fig.~\ref{fig:stage1} shows the intermediate results of SymmFCNet, including warped images, and the completion results after FlowNet, LightNet, and RecNet.
From Fig.~\ref{fig:stage1}, we have the following observations:
(i) FlowNet can correctly align the flip image with the original one, and construct the correspondence between left and right half-faces.
(ii) Although the correspondence can be used to fill in missing pixels within only one half-face, the result suffers from illumination inconsistency, and can be improved via the introduction of LightNet.
(iii) RecNet not only can fill in missing pixels on both of the half-faces, but also is effective in further refining the result of illumination-reweighted warping.

In the second group of experiments, we further assess the {effect of perceptual symmetry loss, FlowNet, and LightNet}.
To this end, we consider five variants of SymmFCNet:
(i) SymmFCNet~(\emph{Full}), (ii) SymmFCNet~(\emph{-S}): removing perceptual symmetry loss, (iii)~SymmFCNet (\emph{-L}): removing LightNet, (iv)~SymmFCNet~(\emph{-GL0}): removing LightNet and applying the predicted flow field only in perceptual symmetry loss, (v)~plain RecNet: removing FlowNet, LightNet and perceptual symmetry loss.
Table~\ref{tab:quan} and Fig.~\ref{fig:ablative} report the quantitative and qualitative results of these variants.
{By removing FlowNet, LightNet and perceptual symmetry loss, plain RecNet only performs on par with Li \etal~\cite{GFC2017CVPR} (Table~\ref{tab:quan}) and is prone to symmetry-inconsistent completion results (Fig.~\ref{fig:ablative} (b)).
%
%In the following, we further assess the effect of FlowNet, perceptual symmetry loss, and LightNet.
}

\textbf{{FlowNet.}}
{
The flow field by FlowNet can be exploited for (i) guiding the completion of missing pixels within only one half-face and (ii) incorporating with $\mathcal{L}_s$ to train RecNet.
Here we only focus on (i) and compare SymmFCNet~(\emph{-L}) and SymmFCNet~(\emph{-GL0}).
By using FlowNet to complete missing pixels within only one half-face, notable gains on PSNR, LPIPS and identity distance can be attained by SymmFCNet~(\emph{-L}) (see Table~\ref{tab:quan}).
From Fig.~\ref{fig:ablative}(c) and Fig.~\ref{fig:ablative}(d), SymmFCNet~(\emph{-GL0}) is still limited in preserving the symmetry consistency of result, while it can be well addressed by SymmFCNet~(\emph{-L}).
%}

\textbf{{Perceptual symmetry loss.}}
{The contribution of perceptual symmetry loss can be assessed by both SymmFCNet~(\emph{-GL0}) vs RecNet and SymmFCNet~(\emph{Full}) vs SymmFCNet~(\emph{-S}).
In comparison to plain RecNet, SymmFCNet~(\emph{-GL0}) can achieve moderate gains on quantitative metrics (see Table~\ref{tab:quan}) and more symmetry consistent results (see Fig.~\ref{fig:ablative}(b)(c)).
It is worth to note that, compared with SymmFCNet~(\emph{-S}), much more gains (\eg, 1.1 dB by PSNR) can be obtained by SymmFCNet~(\emph{Full}).
From Fig.~\ref{fig:ablative} (e) and (f), SymmFCNet~(\emph{Full}) is also able to correct the artifacts and illumination inconsistency produced in the first stage.
Thus, RecNet with perceptual symmetry loss is helpful in filling missing pixels on both of the half-faces and refining the result of illumination-reweighted warping.

}

\textbf{{LightNet}.}
{We further compare SymmFCNet~(\emph{Full}) with SymmFCNet~(\emph{-L}) to assess the contribution of LightNet.
It can be seen that the introduction of LightNet can further improve the quantitative performance (see Table~\ref{tab:quan}) and generate illumination consistent results (see Fig.~\ref{fig:ablative}(d)(f)).
We also note that, RecNet also benefits the correction of illumination inconsistency, and SymmFCNet~(\emph{-L}) attains the second best quantitative performance among the five SymmFCNet varaints.
Even though, from the top image in Fig.~\ref{fig:ablative}(d), illumination inconsistency remains obvious for the result by SymmFCNet~(\emph{-L}), indicating that LightNet is still required and cannot be totally replaced by RecNet.

\section{Conclusion}\label{conclusion}

This work presents a symmetry consistent CNN model, \emph{i.e.}, SymmFCNet, for effective face completion.
In the proposed method, a FlowNet is adopted to construct the correspondence between two half-faces.
The correspondence is then combined with a LightNet for filling in missing pixels within only one half-face, and incorporated with RecNet in the form of perceptual symmetry loss for recovering missing pixels in both of half-faces.
%
%Our full SymmFCNet can be end-to-end learned from training data without correspondence annotations.
%
Extensive experiments show the the effectiveness of SymmFCNet
%in
on generating photo-realistic results with fine details for inpainting rectangular and irregular holes and even real occlusions.
In terms of quantitative metrics, perception quality and user study, our SymmFCNet performs favorably against state-of-the-arts.

{\small
\bibliographystyle{ieee}
\bibliography{egbib}
}
%\vspace{30em}

%\renewcommand\thesection{\Alph{section}}
\renewcommand\thetable{\Alph{table}}
\setcounter{table}{0}

\twocolumn[{\section*{Appendix}}]

Our SymmFCNet consists of three sub-networks, \emph{i.e.,} FlowNet, LightNet and RecNet. FlowNet and LightNet have the same structure except the channel number of the output. Architecture details are shown in Table~\ref{tab:Architecture}. Here, Conv ($d,k,s$) and TransConv ($d,k,s$) denote the convolutional layer and transposed convolutional layer, respectively. $d$, $k$ and $s$ represent output dimension, kernel size and stride, respectively. BN is batch normalization and Concat indicates the concatenation from the $i$-th layer to the ($L-i$)-\emph{th} layer via skip connetion ($L$ is the depth of RecNet). LReLU and DropOut are equipped with parameters of 0.2 and 0.5, respectively. Besides, as for global and part discriminators, the architectures are demonstrated in Table~\ref{tab:discriminator}. PatchGAN is adopted to classify if each $N \times N$ patch in an image is real or fake~\cite{pix2pix2016}.

\begin{strip}
\vspace{1em}
\end{strip}

\begin{strip}
    \begin{tabular}{|c|c| c| c}%
		\hline
		FlowNet & LightNet & RecNet \\
        \hline
        \hline
        \multicolumn{2}{|c|}{Input ($6\times256\times256$)} & Input($6\times256\times256$) \\
        \multicolumn{2}{|c|}{Conv (64, 4, 2), LReLU} & Conv (64, 4, 2), LReLU\\
        \multicolumn{2}{|c|}{Conv (128, 4, 2), BN, LReLU }& Conv (128, 4, 2), BN, LReLU\\
        \multicolumn{2}{|c|}{Conv (256, 4, 2), BN, LReLU }& Conv (256, 4, 2), BN, LReLU\\
        \multicolumn{2}{|c|}{Conv (512, 4, 2), BN, LReLU }& Conv (512, 4, 2), BN, LReLU\\
        \multicolumn{2}{|c|}{Conv (1024, 4, 2), BN, LReLU }& Conv (1024, 4, 2), BN, LReLU\\
        \multicolumn{2}{|c|}{Conv (1024, 4, 2), BN, LReLU }& Conv (1024, 4, 2), BN, LReLU\\
        \multicolumn{2}{|c|}{Conv (1024, 4, 2), BN, LReLU }& Conv (1024, 4, 2), BN, LReLU\\
        \multicolumn{2}{|c|}{Conv (1024, 4, 2), ReLU }& Conv (1024, 4, 2), LReLU\\
        \multicolumn{2}{|c|}{TransConv (1024, 4, 2), BN, ReLU }& TransConv (1024, 4, 2), BN, DropOut, \textbf{Concat}, LReLU\\
        \multicolumn{2}{|c|}{TransConv (1024, 4, 2), BN, ReLU }& TransConv (1024, 4, 2), BN, DropOut, \textbf{Concat}, LReLU\\
        \multicolumn{2}{|c|}{TransConv (1024, 4, 2), BN, ReLU }& TransConv (1024, 4, 2), BN, DropOut, \textbf{Concat}, LReLU\\
        \multicolumn{2}{|c|}{TransConv (512, 4, 2), BN, ReLU }& TransConv (512, 4, 2), BN, \textbf{Concat}, LReLU\\
        \multicolumn{2}{|c|}{TransConv (256, 4, 2), BN, ReLU }& TransConv (256, 4, 2), BN, \textbf{Concat}, LReLU\\
        \multicolumn{2}{|c|}{TransConv (128, 4, 2), BN, ReLU }& TransConv (128, 4, 2), BN, \textbf{Concat}, LReLU\\
        \multicolumn{2}{|c|}{TransConv (64, 4, 2), BN, ReLU }& TransConv (64, 4, 2), BN, \textbf{Concat}, LReLU\\
        \cline{1-2}
        %\hline
        \multicolumn{1}{|c|}{TransConv (2, 4, 2), Tanh} & TransConv (3, 4, 2), ReLU & TransConv (3, 4, 2), Sigmoid\\
        Output($2\times256\times256$) & Output($3\times256\times256$) & Output($3\times256\times256$)\\
        \hline
    \end{tabular}%
    \captionof{table}{Network architecture of SymmFCNet.}
    \label{tab:Architecture}%
\end{strip}

\begin{strip}
    \centering
	\begin{tabular}{ |c| c|}%}
		\hline
		Global Discriminator & Part Discriminator \\
        \hline
        \hline
        Input($3\times256\times256$) & Input($3\times128\times128$) \\
        Conv(64, 4, 2), LReLU & Conv(64, 4, 2), LReLU\\
        Conv(128, 4, 2), BN, LReLU & Conv (128, 4, 2), BN, LReLU\\
        Conv(256, 4, 2), BN, LReLU & Conv (256, 4, 1), BN, LReLU\\
        Conv(512, 4, 1), BN, LReLU & Conv (1, 4, 1), Sigmoid\\
        Conv(1, 4, 1), Sigmoid & Output($1\times30\times30$)\\
        Output($1\times30\times30$) &~\\
        \hline
	\end{tabular}%}
    \captionof{table}{Network architecture of global and part discriminators.}
    \label{tab:discriminator}
\end{strip}

\end{document}